\documentclass[10pt,twocolumn,letterpaper]{article}



\usepackage[pagenumbers]{cvpr} % To force page numbers, e.g. for an arXiv version

\usepackage{svg}
\usepackage{algorithm}% http://ctan.org/pkg
\usepackage{algpseudocode}
\usepackage{amsfonts}
\usepackage{multirow}
\usepackage{amssymb} % for \checkmark
\usepackage{graphicx} % for scaling the table if needed
\usepackage{pifont}% http://ctan.org/pkg/pifont

\usepackage{colortbl}
\definecolor{grey1}{gray}{0.85}
\definecolor{greyG}{rgb}{0.94, 0.88, 0.88}  
\definecolor{greyR}{rgb}{0.88, 0.94, 0.88}
\definecolor{greyB}{rgb}{0.96, 0.96, 0.85}

\definecolor{cvprblue}{rgb}{0.21,0.49,0.74}
\usepackage[breaklinks,colorlinks,allcolors=cvprblue]{hyperref}
\usepackage{multirow}
\usepackage{transparent}
\usepackage{pdflscape}
\title{Exploratory Analysis of Deep Learning Models for Forecasting Meteorological Parameters in the Agricultural Sector}

\author{
Piotr Sikora\\
Faculty of Technical Physics, Information Technology \\ 
and Applied Mathematics \\
Lodz Universtiy of Technlogy\\
{\tt\small 247784@edu.p.lodz.pl}
\and
Sotirios Kontogiannis* \\
MicroComputer Systems Laboratory\\
\small{\url{https://kalipso.math.uoi.gr}}\\
Dept. of Mathematics \\ 
University of Ioannina\\
{\tt\small skontog@uoi.gr} 
}
\begin{document}
\maketitle
\def\thefootnote{*}\footnotetext{Correspondence to: skontog@uoi.gr}
\begin{abstract}
Accurate meteorological forecasting is essential for agricultural planning, irrigation management, and environmental decision support. This study conducts a comparative evaluation of recurrent and hybrid deep learning architectures for multivariate forecasting of reference evapotranspiration ($ET_0$), vapour pressure deficit (VPD), wind speed, and the sine and cosine components of wind direction. The analysis utilizes 134,376 hourly observations from Ioannina, Greece, spanning January 2011 to April 2026, sourced from ERA5 via the OpenMeteo Historical Weather API. Single and multi-layer GRU and LSTM networks are compared with hybrid 1D-CNN-GRU and 1D-CNN-LSTM models for two forecasting tasks: a 24-hour next-day forecast and a 168-hour week-ahead forecast. Performance is evaluated using normalized root mean squared error, the coefficient of determination, and a composite Weighted Quotient Score (WQS). The most effective purely recurrent models are a 64-unit LSTM for the 24-hour horizon, with a WQS of 0.816755, and a 1024-unit GRU for the 168-hour horizon, with a WQS of 0.779465. The hybrid CNN-GRU models achieved the highest overall scores of 0.827535 and 0.782863 for the 24-hour and 168-hour horizons, but with additionally more number of units respectively to LSTM models, while the CNN-LSTM models yield nearly identical results with substantially fewer parameters. Compared to the corresponding recurrent baselines, the hybrid models improve WQS by 1.22--1.63\% at 24 hours and by 0.44--0.45\% at 168 hours, indicating that convolutional feature extraction is more beneficial for short-term forecasting.
\end{abstract}    
\section{Introduction}
\label{sec:intro}

Accurate meteorological forecasting is critical for agricultural planning, irrigation scheduling, crop-water management, and the timely identification of atmospheric conditions that contribute to plant stress. Reference evapotranspiration ($ET_0$) serves as a standardized indicator of atmospheric water demand and is a fundamental parameter for estimating crop-water requirements~\cite{allen1998crop,roy2022et0}. Vapour pressure deficit (VPD) quantifies the difference between saturation and actual vapour pressure, and is closely linked to plant transpiration, water use, and crop response to atmospheric dryness~\cite{vurro2019vpd,elbeltagi2023vpd}. Wind speed and direction are also significant, as they affect the aerodynamic component of evapotranspiration and influence the transport and spatial distribution of heat and moisture. Joint forecasting of these variables can therefore facilitate more informed and timely agricultural decision-making.

Deep learning methods are increasingly utilized in meteorological and environmental forecasting due to their capacity to model nonlinear relationships and capture long-range temporal dependencies. At the global scale, models such as GraphCast have demonstrated the effectiveness of data-driven approaches for medium-range weather prediction~\cite{lam2023graphcast}. At station and regional scales, recurrent neural networks have been employed to predict individual meteorological and agrohydrological variables. Specifically, LSTM and bidirectional LSTM models have been applied to daily and multi-step $ET_0$ forecasting~\cite{roy2022et0}, while recent studies have compared LSTM, temporal convolutional networks, and N-BEATS for reference-evapotranspiration estimation and forecasting~\cite{sarkar2025et0}. Machine-learning techniques have also been investigated for VPD prediction to support agricultural water management~\cite{elbeltagi2023vpd}.

The integration of meteorological prediction with distributed sensing and precision-agriculture platforms enables a direct transition from environmental measurements to operational recommendations. In precision viticulture, the combination of distributed IoT networks, georeferenced sensor data, and deep-learning algorithms has facilitated the detection and prediction of crop stress and disease development~\cite{kontogiannis2026thingsai}. These systems underscore the value of local, high-frequency meteorological information for field-level decision support. For this study, hourly data for Ioannina, Greece, were sourced from the Open-Meteo Historical Weather API~\cite{zippenfenig2023openmeteo}. The underlying reanalysis data originate from ERA5, which offers spatially and temporally consistent meteorological records suitable for long-term environmental analysis~\cite{hersbach2023era5}.

This study conducts an exploratory analysis and a controlled comparative evaluation of recurrent and hybrid neural-network architectures for the joint prediction of $ET_0$, VPD, wind speed, and the sine and cosine components of wind direction components of wind direction. Single- and multi-layer GRU~\cite{chung2014gru} and LSTM~\cite{hochreiter1997lstm,gers2000lstm} architectures were compared with hybrid 1D-CNN-GRU and 1D-CNN-LSTM models~\cite{ige2024cnn1d}. Two forecasting scenarios were considered: a 24-h next-day horizon and a 168-h week-ahead horizon. In contrast to studies focused on a single variable or architecture, the present evaluation examines the combined effects of recurrent cell type, hidden-state size, network depth, convolutional feature extraction, forecasting horizon, and target-variable predictability under a common training and evaluation protocol, as performed in similar scientific reports~\cite{fischer2025comparison, domingos2025exploring}. Performance was quantified using normalized RMSE, $R^2$, and the combined Weighted Quality Score (WQS), providing a consistent basis for identifying suitable architectures for station-scale meteorological forecasting and agricultural decision-support applications.

\section{Dataset and Preprocessing}
\label{sec:data}

\subsection{Data Source}
\begin{figure}
    \centering
    \includegraphics[width=1\linewidth]{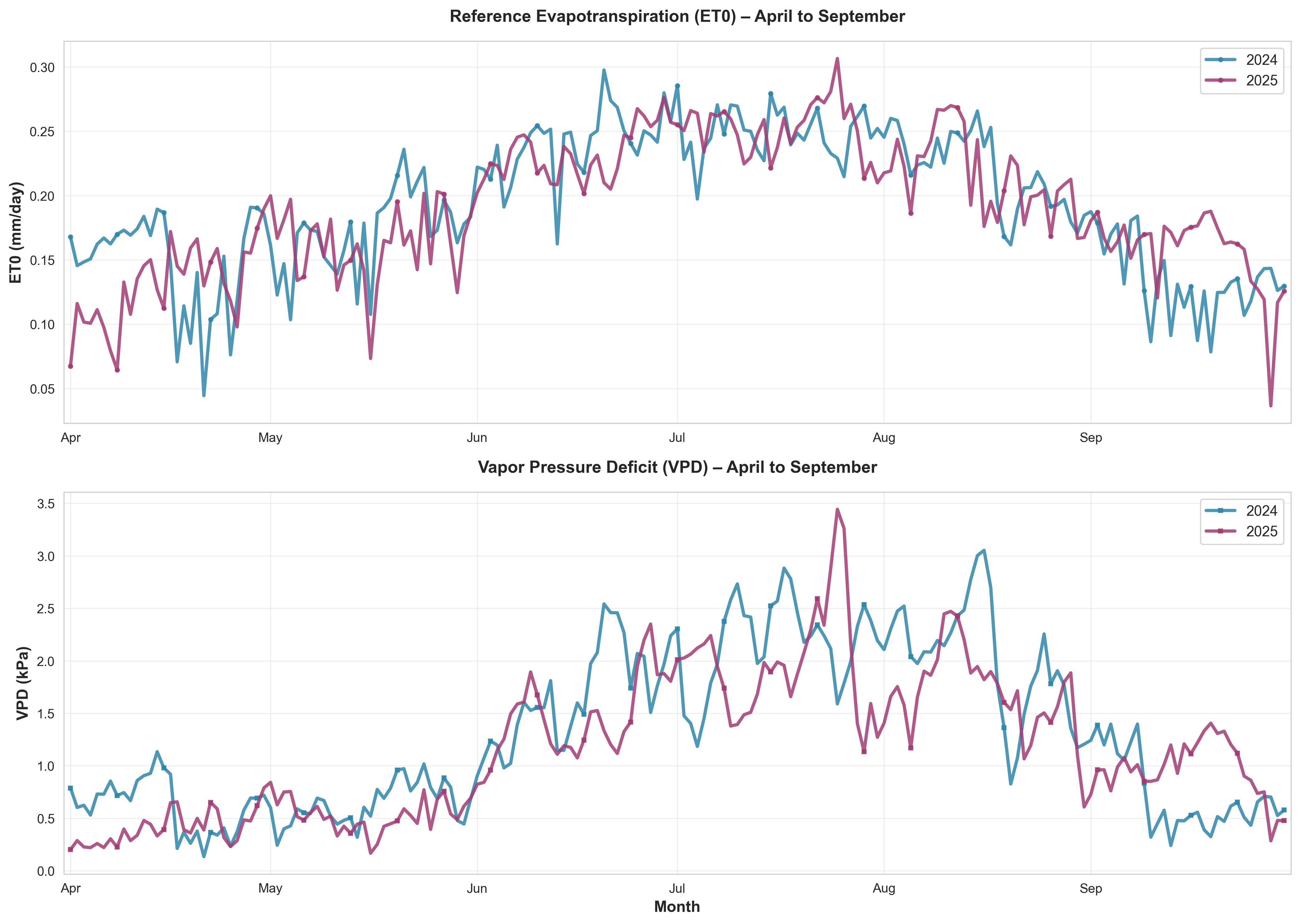}
    \caption{Graphs comparing daily values of Reference Evapotranspiration($ET_0$)) and Vapour Pressure Deficit between 2024 and 2025 from April to September}
    \label{fig:1}
\end{figure}

Hourly meteorological information was collected from the Open-Meteo Weather History API from the start of January~2011 until the end of April~2026. This data comes from ERA5, the fifth generation ECMWF reanalysis for the global climate and weather~\cite{hersbach2020era5} for the pre-existing grid point nearest to the Department of Mathematics of the University of Ioannina. The data was extracted using the Open-Meteo Historical Weather API. For this research, four variables were selected, which were reference evapotranspiration calculated with the FAO method(in $\frac{mm}{h}$), vapour pressure deficit(in kPa), wind speed at height of 10 meters(in $\frac{\text{km}}{\text{h}}$), and wind direction at height of 10 meters(\textdegree{}). The total size of final dataset was 134,376 hourly observations. Figure~\ref{fig:1} illustrates the reference evaportranspiration ($ET_0$) and vapour pressure deficit (VPD) mean values for years 2024--2025 for the area of Ioannina, Epirus, Greece. 

Reference evapotranspiration was obtained directly from the Open-Meteo Historical Weather API using the hourly variable \texttt{et0\_fao\_evapotranspiration}. This variable represents the reference evapotranspiration, $ET_0$, from a hypothetical, actively growing, well-watered grass surface with unlimited soil-water availability. Open-Meteo calculates $ET_0$ according to the FAO-56 Penman--Monteith method, which incorporates air temperature, atmospheric humidity, wind speed, and solar radiation data~\cite{allen1998crop,zippenfenig2023openmeteo}. Each hourly API value is reported in millimeters and represents the evapotranspiration depth accumulated during the preceding hourly interval. Therefore, daily reference evapotranspiration was determined by summing the corresponding hourly values, rather than by calculating their mean.

For an hourly time step, the FAO-56 Penman--Monteith formulation underlying the Open-Meteo reference evapotranspiration variable is calculated using Equation~\eqref{eq:et}.
\begin{equation}
	ET_{0,h} =
	\frac{
		0.408\,\Delta\left(R_n-G\right)
		+
		\gamma
		\left(
		\frac{37}{T+273}
		\right)
		u_2\,VPD
	}{
		\Delta+\gamma\left(1+0.34u_2\right)
	}
	\label{eq:et}
\end{equation}

where $ET_{0,h}$ is the reference evapotranspiration accumulated during the hour in $\mathrm{mm}$, $R_n$ is the net radiation at the reference surface in $\mathrm{MJ\,m^{-2}\,h^{-1}}$, $G$ is the soil heat-flux density in
$\mathrm{MJ\,m^{-2}\,h^{-1}}$, $T$ is the hourly air temperature at a height of $2$~m in \textdegree C, and $u_2$ is the wind speed in m/sec. Moreover, $e_s$ and $e_a$ are the saturation and actual vapour pressures in $\mathrm{kPa}$, respectively, $\Delta$ is the slope of the saturation vapour-pressure curve in $\mathrm{kPa\,^{\circ}C^{-1}}$, and $\gamma$ is the psychrometric constant in $\mathrm{kPa\,^{\circ}C^{-1}}$. The first term in the numerator represents the energy available for evapotranspiration, whereas the second term represents the aerodynamic contribution associated with wind speed and atmospheric vapour demand~\cite{allen1998crop}.

Vapour pressure deficit (VPD) was retrieved from Open-Meteo using the hourly variable \texttt{vapour\_pressure\_deficit} and is expressed in $\mathrm{kPa}$. In contrast to hourly $ET_0$, VPD is reported as an instantaneous value for the indicated time. VPD represents the difference between the saturation vapour pressure at the prevailing air temperature and the actual vapour pressure of the atmosphere. Using air temperature $T$ in \textdegree C and relative humidity $RH$ in percent, VPD can be calculated using Equation~\eqref{eq:vpd}.
\begin{align}
	VPD
	&=
	e_s(T)-e_a
	=
	e_s(T)
	\left(
	1-\frac{RH}{100}
	\right),
	\label{eq:vpd}\\
	e_s(T)
	&=
	0.6108
	\exp\left(
	\frac{17.27T}{T+237.3}
	\right),
	\notag\\
	e_a
	&=
	\frac{RH}{100}e_s(T),
	\notag\\
	\Delta
	&=
	\frac{4098\,e_s(T)}
	{\left(T+237.3\right)^2},
	\notag\\
	\gamma
	&=
	0.000665\,P.
	\notag
\end{align}

Higher VPD values indicate greater atmospheric demand for water, while values near zero suggest that the air is approaching saturation. Since VPD is included in the aerodynamic component of Equation~\eqref{eq:et}, an increase in VPD generally enhances the atmospheric driving force for evaporation and plant transpiration~\cite{allen1998crop}.

\subsection{Data Preprocessing}
Data preprocessing initiates with the alignment of hourly records to a regular 1-hour grid. To solve the problem of missing values, the method of forward-fill imputation was implemented in the preprocessing pipeline, but no missing values were found in the dataset. The wind direction, which was originally measured in degrees from 0\textdegree{} to 360\textdegree{}, had to be transformed, due to a discontinuity at 0\textdegree{}/360\textdegree{}, using Equations ~\eqref{eq:wind1},~\eqref{eq:wind2}. Therefore, the variable was replaced by its sine and cosine projections, expanding the feature set from four raw variables to five model input values of the transformed direction vector.

\begin{equation}
  w_{\sin} = \sin\!\left(\frac{2\pi\,\theta}{360}\right)
  \label{eq:wind1}
\end{equation}
\begin{equation}
  w_{\cos} = \cos\!\left(\frac{2\pi\,\theta}{360}\right)
  \label{eq:wind2}
\end{equation}

Then the normalization of all variables follows to a scale between 0 and 1. This was achieved by using per variable independent min-max scalers of the Python scikit-learn API~\cite{scikit-learn}, that were calculated on the whole time series dataset. The per variable instantiated scaler objects were saved for the future cases of inference. The normalized time series was then divided into overlapping sliding windows using a stride sampling method. Two different window configurations were created for the experiment. The first is short-term windows which have 72 hours of input lookback, 24 hours of forecasting horizon, and 12 hours of stride. The second is long-term windows which have 336 hours of input lookback , 168 hours of forecasting horizon and 48 hours of stride. For both configurations, the total set of windows was shuffled and divided with a fixed random seed into train part of 70\%, validation part of 15\% and test part of 15\%. This strategy of sampling is useful because it maintains the temporal connection inside each window, and it also allows for the model to experience many different meteorological regimes. Table~\ref{tab1} presents the sliding window parameters and train validation and test sizes used from the OpenMeteo meteorological data for the area of Ioannina, Epirus, Greece for the years 2011--2026 of hourly resolution.

\begin{table}[ht]
\centering
\caption{Sliding windows count and train, validation and test data sizes for daily and weekly forecasts. The dataset has been colllected using the OpenMeteo~\cite{zippenfenig2023openmeteo} meteorological hourly data from the area of Ioannina, Epirus, Greece}
\label{tab1}
\resizebox{1\linewidth}{!}{
\begin{tabular}{r r r c c c}
\hline
\textbf{Horizon} & \textbf{Lookback} & \textbf{Stride} & \textbf{Train set size} & \textbf{Validation set size} & \textbf{Test set size} \\
\hline \hline
24 h   & 72 h  & 12 & 7833 & 1678 & 1680 \\
\hline
168 h  & 336 h & 48 & 1952 & 418 & 420 \\
\hline
\end{tabular}
}
\end{table}

\section{Experimental Scenario}
\label{sec:setup}

Two forecasting scenario cases have been evaluated. In the short-term scenario, a sequence of 72 hourly observations, representing the previous three days, served as input to predict the 24 hourly values of the subsequent day. In the weekly scenario, the input window comprised 336 hours, equivalent to two full weeks, and the prediction horizon was 168 hours, corresponding to the following week. These horizons were chosen to facilitate both next-day forecasting and medium-term operational planning. The selected input lengths enable the models to capture recurrent temporal patterns, such as daily and weekly periodicities.

Training samples were generated using sliding windows with a stride of 12 hours for the short-term scenario and 48 hours for the weekly scenario. Samples were processed in mini-batches of 32. For each batch, the input tensor $\mathbf{X}$ had shape $(32, L, 5)$, where $L$ represents the scenario-specific lookback length and the final dimension corresponds to the five meteorological variables. The target tensor $\mathbf{y}$ had shape $(32, H, 5)$, where $H$ denotes the prediction horizon. Specifically, $L=72$ and $H=24$ for the short-term scenario, and $L=336$ and $H=168$ for the weekly scenario.

All models put to the test, describen in section~\ref{sec:models}, employed dropout regularization with a probability of $p=0.2$ to mitigate overfitting. Training utilized the Adam optimizer~\cite{kingma2017adam}, with an initial learning rate of $10^{-3}$ and a weight decay coefficient of $10^{-4}$. A learning rate scheduler reduced the learning rate by a factor of 0.5 if the validation loss, measured by mean squared error, did not improve for five consecutive epochs. The maximum number of training epochs was set to 200, and gradient clipping limited the maximum $\ell_2$ norm to 1.0. Early stopping was implemented with a patience of 10 epochs and a minimum improvement threshold of $10^{-5}$.

\subsection{Evaluation Metrics}
\label{sec:metrics}

Model performance was evaluated using the normalized root mean squared error (nRMSE-RMSE), the coefficient of determination ($R^2$), and a composite Weighted Quality Score (WQS). The RMSE was calculated using Equation~\eqref{eq:nrmse}, by normalizing the root mean square error squared error by the range of ground-truth values in the test partition.

\begin{equation}
	\mathrm{RMSE} =
	\frac{
		\displaystyle
		\sqrt{\frac{1}{N}\sum_{i=1}^{N}
			\left(y_i-\hat{y}_i\right)^2}
	}{
		y_{\max}-y_{\min}
	},
	\label{eq:nrmse}
\end{equation}
\noindent where $N$ is the number of evaluated observations, $y_i$ and $\hat{y}_i$ denote the observed and predicted values, respectively, and $y_{\max}$ and $y_{\min}$ is the maximum and minimum ground-truth values in the test. This normalization yields a dimensionless error metric, which facilitates comparison across forecasting horizons and target variables. Lower RMSE values correspond to improved predictive performance, with a value of zero indicating perfect predictions. The coefficient of determination was computed using Equation~\eqref{eq:r2}.

\begin{equation}
	R^2 =
	1-
	\frac{
		\displaystyle\sum_{i=1}^{N}
		\left(y_i-\hat{y}_i\right)^2
	}{
		\displaystyle\sum_{i=1}^{N}
		\left(y_i-\bar{y}\right)^2
	},
	\label{eq:r2}
\end{equation}
\noindent where $\bar{y}$ is the mean of the observed values. An $R^2$ value of one indicates a perfect fit, while a value of zero reflects performance equivalent to predicting the sample mean. Negative values indicate that the model performs worse than the sample-mean baseline. In the implemented evaluation procedure, $R^2$ values were restricted to the interval $[0,1]$. As a result, models with severely poor predictive performance received an $R^2$ score of zero. To provide a unified performance indicator that integrates prediction error and explained variance, the Weighted Quality Score was defined using Equation~\eqref{eq:wqs}, inspired by the WQS definition in~\cite{tsolaki26}.
\begin{equation}
	\mathrm{WQS}
	=
	\alpha\left(1-\mathrm{RMSE}\right)
	+
	(1-\alpha)R^2,
	\label{eq:wqs}
\end{equation}
where $\alpha=0.8$. Thus, 80\% of the composite score was assigned to normalized predictive accuracy and 20\% to explained variance. Greater WQS values indicate superior overall forecasting performance. Since RMSE $\geq 0$ and the implemented $R^2$ values are restricted to $[0,1]$, a WQS value greater than one would indicate an error in the metric normalization or implementation.

The three metrics were first calculated over the complete model output by aggregating all test samples, prediction steps, and target variables. Additionally, a separate WQS value was calculated for each target variable. This variable-wise analysis quantifies how accurately each meteorological quantity was predicted when all five variables were jointly supplied as model inputs.
\section{Model Architectures}\label{sec:models}

\begin{figure}[ht]
	\centering
	\includegraphics[width=1\linewidth]{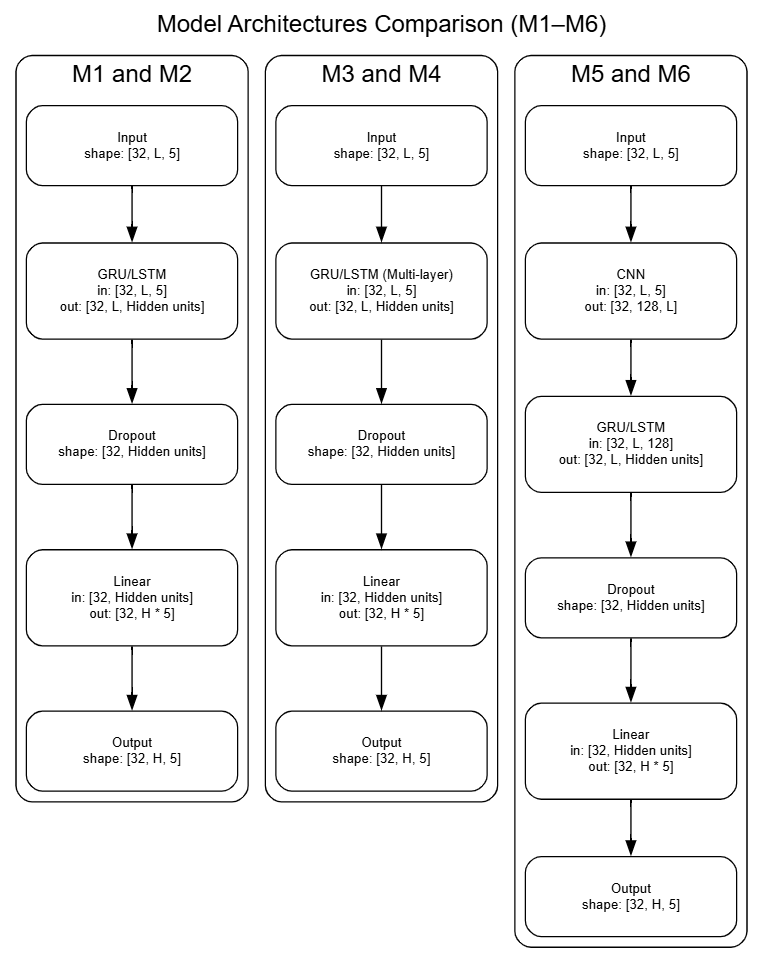}
	\caption{Comparison of the six forecasting architectures. M1 and M2 are single-layer GRU and LSTM models, respectively. M3 and M4 are multi-layer GRU and LSTM models. M5 and M6 combine a one-dimensional convolutional feature extractor with a GRU and an LSTM, respectively. Here, $L$ denotes the input-sequence length and $H$ denotes the forecasting horizon.}
\label{fig:2}
\end{figure}

Six neural network model architectures were evaluated, including four recurrent models (M1 to M4) and two hybrid convolutional-recurrent models (M5 and M6). The overall structures of these models are shown in Figure~\ref{fig:2}. Each model receives an input tensor $\mathbf{X}\in\mathbb{R}^{B\times L\times 5}$, where $B$ denotes the mini-batch size.
$L$ represents the scenario-specific lookback length, and the final dimension contains the five meteorological variables. For a complete mini-batch, $B=32$. All architectures generate a multi-step output tensor $\hat{\mathbf{Y}}\in\mathbb{R}^{B\times H\times 5}$, where $H=24$ for the short-term scenario and $H=168$ for the weekly scenario.

Models M1 and M2 are single-layer recurrent architectures based on the Gated Recurrent Unit (GRU) and Long Short-Term Memory (LSTM) cell, respectively. GRU and LSTM networks are designed to capture long-range temporal dependencies and mitigate the vanishing-gradient problem associated with conventional recurrent neural networks~\cite{hochreiter1997lstm,chung2014gru}. The recurrent layer processes the entire input sequence, and its final hidden representation is passed through a dropout layer and a fully connected projection layer. The linear layer outputs $H\times5$ values, which are then reshaped into the required output tensor of size $(B,H,5)$. Hidden-state sizes of 10, 64, 256, and 1024 units were evaluated for both models.

Models M3 and M4 extend the recurrent structure by stacking multiple GRU and LSTM layers, respectively. In these architectures, the output sequence of one recurrent layer serves as the input to the subsequent layer.
This design enables the network to learn progressively more abstract temporal representations. Dropout with probability $p=0.2$ was applied between successive recurrent layers and before the final projection head~\cite{srivastava2014dropout}. Two depth-width configurations were investigated: (1) four recurrent layers with 64 hidden units and (2) ten recurrent layers with 1024 hidden units.
The final hidden state of the uppermost recurrent layer was mapped to the $H\times5$ forecasting vector and subsequently reshaped into $(B,H,5)$.

Models M5 and M6 are hybrid CNN-RNN architectures that integrate local temporal feature extraction with recurrent sequence modeling. M5 employs a GRU following the convolutional component, while M6 utilizes an LSTM. The input tensor is initially transposed from $(B,L,5)$ to $(B,5,L)$, as required by the PyTorch one-dimensional convolution implementation~\cite{paszke2019pytorch}. The convolutional stack expands the five input channels into a 128-channel representation, preserving the sequence length by using a kernel size of three and same padding.
tensor therefore has shape $(B,128,L)$, as shown in Figure~\ref{fig:2}.

The convolutional feature extractor consists of two successive one-dimensional convolutional blocks. Each block combines a
Conv1D operation, batch normalization, a rectified linear unit activation, and dropout with $p=0.2$. Batch normalization stabilizes the distribution of intermediate activations and can improve optimization during training~\cite{ioffe2015batchnorm}. One-dimensional convolutions capture short-term patterns and interactions among adjacent hourly observations~\cite{ige2024cnn1d}. Before entering the recurrent layer, the convolutional output is transposed back to $(B,L,128)$ so that the temporal dimension is interpreted as the sequence axis.

The recurrent component of each hybrid model integrates the local patterns identified by the CNN across the entire lookback period. The final GRU or LSTM hidden representation is passed through dropout and a linear projection layer, consistent with the output strategy used by M1 to M4. For the short-term forecasting task, M5 used 256 hidden units, while M6 used 64 hidden units. For the weekly task, M5 used 1024 hidden units and M6 maintained a hidden size of 64. These configurations were chosen to assess whether the convolutional feature-extraction stage could reduce the recurrent capacity required for accurate multi-step forecasting.

All architectures were implemented in PyTorch~\cite{paszke2019pytorch}. A common output projection, dropout rate, optimization procedure, batch size, and stopping criteria were used across all models.
rate, optimization procedure, batch size, and stopping criteria ensured that differences in forecasting performance could be attributed primarily to the architectural choices. In particular, the comparison investigates the effects of recurrent cell type, hidden-state dimensionality, recurrent depth, and the inclusion of convolutional feature extraction.

The training and validation results in Table~\ref{tab:train_res} indicate that increasing model size did not consistently improve generalization. For the 24-hour horizon, the lowest validation losses were achieved by the hybrid models M6 and M5. The validation losses for M6 and M5 were 0.025300 and 0.025396, respectively. For the 168-hour horizon, M5 achieved the lowest validation loss of 0.034103, closely followed by M6 with 0.034180. These results suggest that convolutional preprocessing provided useful local temporal features prior to recurrent modeling, particularly for the longer forecasting horizon.

\begin{table}[ht]
\centering
\caption{Training and validation results for models across both horizons.}
\label{tab:train_res}
\resizebox{1\linewidth}{!}{
\begin{tabular}{l r r r r r r r r}
\hline
\textbf{Model} & \textbf{Hidden} & \textbf{Layers} & \textbf{Horizon} &
\textbf{Last Train Loss} & \textbf{Best Val Loss} &
\textbf{Epochs} & \textbf{Time (s)} & \textbf{Params} \\
\hline\hline

\multirow{8}{*}{\textbf{M1}}
& \multirow{2}{*}{10}   & \multirow{2}{*}{1} & 24  & 0.032769 & 0.030017 & 77  & 217  & 1830 \\
&                       &                    & 168 & 0.041036 & 0.039308 & 91  & 135  & 9750 \\
& \multirow{2}{*}{64}   & \multirow{2}{*}{1} & 24  & 0.029620 & 0.028776 & 43  & 137  & 21432 \\
&                       &                    & 168 & 0.036533 & 0.035588 & 48  & 85   & 68232 \\
& \multirow{2}{*}{256}  & \multirow{2}{*}{1} & 24  & 0.028567 & 0.028155 & 85  & 255  & 232824 \\
&                       &                    & 168 & 0.039794 & 0.035990 & 14  & 50   & 417864 \\
& \multirow{2}{*}{1024} & \multirow{2}{*}{1} & 24  & 0.029232 & 0.028899 & 50  & 736  & 3290232 \\
&                       &                    & 168 & 0.034590 & 0.034746 & 96  & 1424 & 4028232 \\
\hline

\multirow{8}{*}{\textbf{M2}}
& \multirow{2}{*}{10}   & \multirow{2}{*}{1} & 24  & 0.032510 & 0.029826 & 76  & 215  & 2000 \\
&                       &                    & 168 & 0.037725 & 0.035981 & 157 & 228  & 9920 \\
& \multirow{2}{*}{64}   & \multirow{2}{*}{1} & 24  & 0.028409 & 0.027452 & 107 & 303  & 25976 \\
&                       &                    & 168 & 0.035657 & 0.035035 & 75  & 124  & 72776 \\
& \multirow{2}{*}{256}  & \multirow{2}{*}{1} & 24  & 0.028060 & 0.027528 & 96  & 287  & 300152 \\
&                       &                    & 168 & 0.036947 & 0.035282 & 45  & 140  & 485192 \\
& \multirow{2}{*}{1024} & \multirow{2}{*}{1} & 24  & 0.028181 & 0.027666 & 95  & 1704 & 4345976 \\
&                       &                    & 168 & 0.036326 & 0.036114 & 29  & 695  & 5083976 \\
\hline

\multirow{4}{*}{\textbf{M3}}
& \multirow{2}{*}{64}   & \multirow{2}{*}{4} & 24  & 0.031212 & 0.030092 & 38 & 300 & 96312 \\
&                       &                    & 168 & 0.036321 & 0.035610 & 61 & 317 & 143112 \\
& \multirow{2}{*}{1024} & \multirow{2}{*}{10} & 24 & 0.061553 & 0.061604 & 10 & 3252 & 59968632 \\
&                       &                    & 168 & 0.049241 & 0.049166 & 5  & 2742 & 60706632 \\
\hline

\multirow{4}{*}{\textbf{M4}}
& \multirow{2}{*}{64}   & \multirow{2}{*}{4} & 24  & 0.031779 & 0.031065 & 97  & 684  & 125816 \\
&                       &                    & 168 & 0.036487 & 0.035797 & 125 & 602  & 172616 \\
& \multirow{2}{*}{1024} & \multirow{2}{*}{10} & 24  & 0.061477 & 0.061673 & 12  & 4471 & 79917176 \\
&                       &                    & 168 & 0.049158 & 0.049263 & 1   & 2520 & 80655176 \\
\hline

\multirow{2}{*}{\textbf{M5}}
&                  256  &                  1 & 24  & 0.026056 & 0.025396 & 96  & 348  & 353400 \\
&                 1024  &                  1 & 168 & 0.033948 & 0.034103 & 69  & 1144 & 4432200 \\
\hline

\multirow{2}{*}{\textbf{M6}}
& \multirow{2}{*}{64}   &                  1 & 24  & 0.026620 & 0.025300 & 73  & 291 & 83576 \\
&                       &                  1 & 168 & 0.034741 & 0.034180 & 80  & 156 & 130376 \\
\hline

\end{tabular}
}
\end{table}

\begin{figure}
    \centering
    \includegraphics[width=1\linewidth]{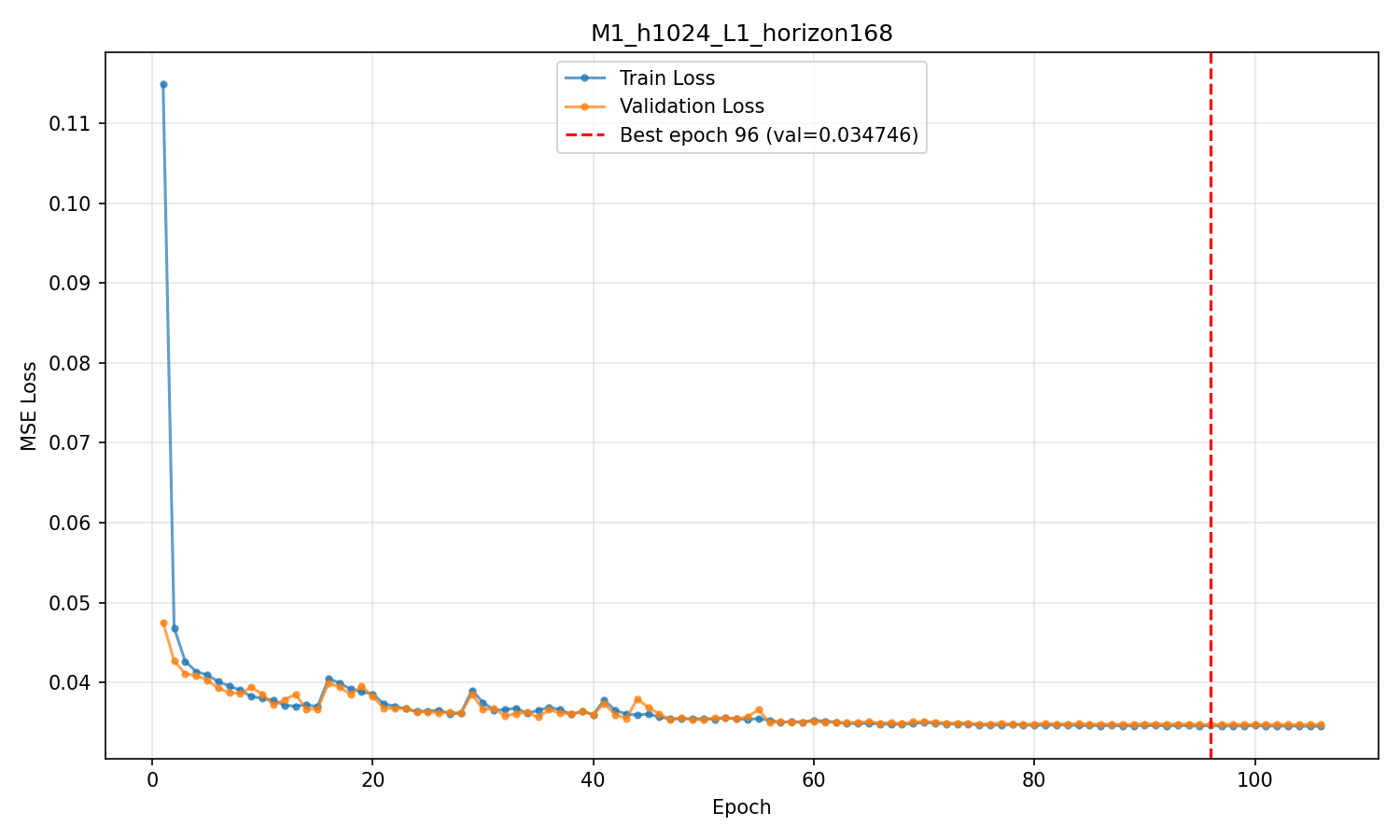}
    \caption{Training and validation loss curves for the single-layer M1 GRU model with 1024 hidden units and a forecasting horizon of 168~h.}
    \label{fig:3}
\end{figure}

The most complex M3 and M4 configurations, each comprising ten recurrent layers and 1024 hidden units, required approximately 60 to 81 million trainable parameters. However, these models exhibited substantially higher validation losses compared to smaller architectures and required significantly longer training times. These results suggest that excessive model depth and parameterization led to optimization challenges and overfitting, without yielding improvements in predictive accuracy. In contrast, the hybrid models, particularly M6, achieved superior validation performance with considerably fewer parameters. Figure~\ref{fig:3} illustrates a representative optimization trajectory for the M1 model with 1024 hidden units in the weekly forecasting scenario. The consistent reduction in both training and validation loss demonstrates stable convergence, and the close alignment of the two curves indicates effective generalization. The use of a learning rate scheduler and early-stopping mechanism prevented unnecessary training once the validation loss plateaued.

\section{Results and Discussion}
\label{sec:results}

\subsection{Recurrent Models (GRU/LSTM)}

For the 24-h forecasting horizon, the single-layer recurrent architectures consistently outperformed their deeper counterparts. The best overall configuration was the single-layer LSTM model M2 with 64 hidden units,
which achieved an RMSE of 0.138704, an $R^2$ of 0.638589, and a WQS of 0.816755. The best GRU configuration was M1 with 256 hidden units, which obtained a WQS of 0.814284. Thus, the best LSTM exceeded the best GRU by only 0.002471 WQS units, corresponding to an approximately 0.3\% relative improvement. This small difference indicates that both recurrent cell types were similarly effective for next-day forecasting.

Increasing recurrent depth did not improve predictive performance. The four-layer GRU and LSTM models produced WQS values of 0.801306 and 0.797921, respectively, which were approximately 2\% lower than those of the best single-layer configurations. The ten-layer models exhibited a much larger degradation, with WQS values close to 0.6235 and $R^2$ values of approximately 0.033. These results suggest that the additional depth made the models more difficult to optimise and did not provide useful additional temporal representations. Moreover, increasing the hidden-state size beyond 64 units produced only marginal differences among the single-layer models, indicating diminishing returns from increased recurrent capacity. The complete results for the 24-h horizon are reported in Table~\ref{tab:recurrent_24_overall}.

\begin{table}[ht]
\centering
\caption{Test-set performance of the recurrent models (GRU/LSTM) for the 24-h forecasting horizon.}
\label{tab:recurrent_24_overall}
\resizebox{1\linewidth}{!}{
\begin{tabular}{l r r r r r r r}
\hline
\textbf{Model} & \textbf{Hidden} & \textbf{Layers} &
\textbf{RMSE}&
\textbf{R$^2$} &
\textbf{WQS} \\

\hline \hline
\multirow{6}{*}{\textbf{GRU (M1 and M3)}} 
& \textbf{256}	& \textbf{1}  & \textbf{0.140169} & \textbf{0.632098} &	\textbf{0.814284}  \\
& 64	& 1  & 0.141941 & 0.62204  &	0.810855 \\
& 1024	& 1  & 0.142097 & 0.621017 &	0.810526 \\
& 64	& 4  & 0.146326 & 0.591835 &	0.801306 \\
& 10	& 1  & 0.147371 & 0.589412 &	0.799985 \\
& 1024	& 10 & 0.228924 & 0.033424 &	0.623545 \\
\hline
\multirow{6}{*}{\textbf{LSTM (M2 and M4)}} 
& \textbf{64} &	\textbf{1} &	\textbf{0.138704} &	\textbf{0.638589} &	\textbf{0.816755} \\
& 256 &	1 &	0.138958 &	0.637709 &	0.816376 \\
& 1024 &	1 &	0.139225 &	0.636588 &	0.815938 \\
& 10 &	1 &	0.146872 &	0.594949 &	0.801492 \\
& 64 &	4 &	0.148302 &	0.582815 &	0.797921 \\
& 1024 &	10 &	0.228997 &	0.033340 &	0.623471 \\
\hline
\end{tabular}
}
\end{table}

For the longer 168-h horizon, the overall performance decreased, as expected, because the models were required to generate forecasts over an entire week. Nevertheless, the same general architectural pattern was observed: a) single-layer models remained the strongest configurations. The best result was obtained by the single-layer GRU model M1 with 1024 hidden units, which achieved an RMSE of 0.158622, an $R^2$ of 0.531813, and a WQS of 0.779465. The single-layer LSTM model M2 with 64 hidden units followed closely, with a WQS of 0.778542. The difference between the two best configurations was only 0.000923 WQS units, confirming that GRU and LSTM cells provided nearly equivalent performance for weekly forecasting.

The four-layer models produced intermediate results and remained relatively close to the strongest single-layer configurations. In contrast, the ten-layer GRU and LSTM models achieved WQS values of 0.699981 and
0.698800, respectively, corresponding to approximately 89.8\% of the WQS obtained by the best model. Their $R^2$ values were also substantially lower, falling below 0.29. Although the GRU model with 1024 hidden units
provided the best weekly result, the improvements over the smaller single-layer configurations were limited. Consequently, the results again indicate that increasing model width or depth does not necessarily improve generalisation. The complete 168-h results are presented in Table~\ref{tab:recurrent_168_overall}.

\begin{table}[ht]
\centering
\caption{Test-set performance of the recurrent models for the 168-h forecasting horizon.}
\label{tab:recurrent_168_overall}
\resizebox{1\linewidth}{!}{
\begin{tabular}{l r r r r r r r}
\hline
\textbf{Model} & \textbf{Hidden} & \textbf{Layers} &
\textbf{RMSE}&
\textbf{R$^2$} &
\textbf{WQS} \\

\hline \hline
\multirow{6}{*}{\textbf{GRU (M1 and M3)}} 
& \textbf{1024} &	\textbf{1} &	\textbf{0.158622} & 	\textbf{0.531813} &	\textbf{0.779465} \\
& 64 &	1 &	0.160930 & 	0.515367 &	0.774329 \\
& 64 &	4 &	0.162168 & 	0.513046 &	0.772875 \\
& 256 &	1 &	0.161760 & 	0.510619 &	0.772715 \\
& 10 &	1 &	0.167308 & 	0.482439 &	0.762642 \\
& 1024 &	10 &	0.197272 & 	0.288991 &	0.699981 \\
\hline
\multirow{6}{*}{\textbf{LSTM(M2 and M4)}} 
& \textbf{64} &	\textbf{1} &	\textbf{0.159163} &	\textbf{0.529363} &	\textbf{0.778542} \\
& 256 &	1 &	0.15976 &	0.526692 &	0.777531 \\
& 64 &	4 &	0.160332 &	0.521532 &	0.776041 \\
& 10 &	1 &	0.161405 &	0.515974 &	0.774071 \\
& 1024 &	1 &	0.161843 &	0.51533 &	0.773592 \\
& 1024 &	10 &	0.197703 &	0.284810 &	0.698800 \\

\hline
\end{tabular}
}
\end{table}

The variable-specific evaluation for the 24-h horizon revealed substantial differences in predictability among the five target variables. Reference evapotranspiration ($ET_0$) and vapour pressure deficit exhibited the
highest WQS values, exceeding 0.94 for the strongest LSTM configurations. Wind speed was more difficult to predict, with the best WQS remaining slightly below 0.80. The sine and cosine components of wind direction were the least predictable variables, with maximum WQS values of approximately 0.699 and 0.697, respectively.

The single-layer LSTM models achieved the strongest variable-specific performance in most cases. M2 with 64 hidden units produced the best $ET_0$ and $w_{\sin}$ scores, whereas M2 with 256 hidden units achieved the highest VPD and wind-speed scores. Differences among the 64-, 256-, and 1024-unit LSTM configurations were, however, very small. The
four-layer models showed moderate degradation, while the ten-layer models performed considerably worse for all variables. The per-variable results for the 24-h horizon are summarised in Table~\ref{tab:recurrent_24_variable}.

\begin{table}[ht]
\centering
\caption{Variable-specific WQS values of the recurrent models for the 24-h forecasting horizon.}
\label{tab:recurrent_24_variable}
\resizebox{1\linewidth}{!}{
\begin{tabular}{l r r r r r r r}
\hline
\textbf{Model} & \textbf{Hidden} & \textbf{Layers} &
\textbf{WQS (ET0)} &
\textbf{WQS (VPD)} &
\textbf{WQS (Wind speed)} &
\textbf{WQS ($w_{sin}$)} &
\textbf{WQS ($w_{cos}$)} \\

\hline \hline
\multirow{4}{*}{\textbf{M1}} 
& 10	& 1	& 0.938591 &	0.915180 &	0.776789 &	0.686912 &	0.682455 \\
& 64	& 1	& 0.946864 &	0.933217 &	0.794159 &	0.692238 &	0.687796 \\
& 256	& 1	& 0.947749 &	0.937741 &	0.798892 &	0.695315 &	0.691725 \\
& 1024	& 1	& 0.946395 &	0.936503 &	0.792462 &	0.691440 &	0.685830 \\
\hline
\multirow{4}{*}{\textbf{M2}} 
& 10	& 1	& 0.936597 &	0.917702 &	0.781657 &	0.687063 &	0.684440 \\
& 64	& 1	& \textbf{0.948005} &	0.940510 &	0.799189 &	\textbf{0.699428} &	0.696642 \\
& 256	& 1	& 0.947594 &	\textbf{0.940687} &	\textbf{0.799244} &	0.698056 &	0.696297 \\
& 1024	& 1	& 0.946884 &	0.939765 &	0.798997 &	0.697749 &	0.696294 \\
\hline
\multirow{2}{*}{\textbf{M3}} 
& 64	& 4	& 0.940988 &	0.929525 &	0.769126 &	0.686876 &	0.680016 \\
& 1024	& 10	& 0.666084 &	0.695625 &	0.705905 &	0.517529 &	0.532584 \\
\hline
\multirow{2}{*}{\textbf{M4}} 
& 64	& 4	& 0.940288 &	0.927118 &	0.768373 &	0.678431 &	0.675396 \\
& 1024	& 10	& 0.666204 &	0.695572 &	0.706510 &	0.516050 &	0.533018 \\
\hline
\end{tabular}
}
\end{table}

The same ordering of variable predictability was observed for the 168-h forecasting horizon (weekly forecasts), although the WQS values generally decreased as the prediction interval increased. Reference evapotranspiration remained the most predictable variable, with a maximum WQS of 0.932572, followed by VPD, whose maximum WQS was 0.899691. Wind speed achieved a maximum score of 0.778015, while the transformed wind-direction components remained the most challenging targets.

For the weekly horizon, M1 with 1024 hidden units achieved the best $ET_0$, wind-speed, $w_{\sin}$, and $w_{\cos}$ scores. M2 with 256 hidden units obtained the highest VPD score. Nevertheless, the differences among the strongest single-layer configurations remained small, particularly for wind speed and the wind-direction components. The cosine component was consistently more difficult to predict than the sine component, although the size of this difference varied across model configurations.

The four-layer recurrent models remained competitive for some variables. For example, the four-layer LSTM obtained an $ET_0$ WQS of 0.932475, which was almost identical to the best value of 0.932572. In contrast, the ten-layer models showed substantial reductions for $ET_0$, VPD, and the wind-direction components. This finding suggests that excessive recurrent depth particularly affected variables with stronger predictable temporal structure, while offering no meaningful advantage for inherently more irregular wind-related variables. The variable-specific weekly results are presented in Table~\ref{tab:recurrent_168_variable}.

\begin{table}[ht]
\centering
\caption{Variable-specific WQS values of the recurrent models for the 168-h forecasting horizon.}
\label{tab:recurrent_168_variable}
\resizebox{1\linewidth}{!}{
\begin{tabular}{l r r r r r r r}
\hline
\textbf{Model} & \textbf{Hidden} & \textbf{Layers} &
\textbf{WQS (ET0)} &
\textbf{WQS (VPD)} &
\textbf{WQS (Wind speed)} &
\textbf{WQS ($w_{sin}$)} &
\textbf{WQS ($w_{cos}$)} \\

\hline \hline
\multirow{4}{*}{\textbf{M1}} 
&	10 &	1 &	0.92601 &	0.879113 &	0.771133 &	0.645857 &	0.591095 \\
&	64 &	1 &	0.930696 &	0.886798 &	0.771243 &	0.652832 &	0.630076 \\
&	256 &	1 &	0.929193 &	0.886533 &	0.769065 &	0.651416 &	0.627370 \\
&	1024 &	1 &	\textbf{0.932572} &	0.898388 &	\textbf{0.778015} &	\textbf{0.655488} &	\textbf{0.632864} \\
\hline
\multirow{4}{*}{\textbf{M2}} 
&	10 &	1 &	0.928105 &	0.88993 &	0.774109 &	0.651415 &	0.626795 \\
&	64 &	1 &	0.930347 &	0.898303 &	0.777015 &	0.654852 &	0.632194 \\
&	256 &	1 &	0.927232 &	\textbf{0.899691} &	0.775148 &	0.654293 &	0.631290 \\
&	1024 &	1 &	0.925070 &	0.893813 &	0.773682 &	0.65013 &	0.625266 \\
\hline
\multirow{2}{*}{\textbf{M3}} 
&	64 &	4 &	0.926165 &	0.887515 &	0.77579 &	0.650982 &	0.623922 \\
&	1024 &	10 &	0.828012 &	0.760069 &	0.757285 &	0.615434 &	0.539104 \\
\hline
\multirow{2}{*}{\textbf{M4}} 
&	64 &	4 &	0.932475 &	0.894495 &	0.775829 &	0.653578 &	0.623828 \\
&	1024 &	10 &	0.827203 &	0.755133 &	0.757716 &	0.614853 &	0.539093 \\
\hline
\end{tabular}
}
\end{table}

\subsection{Hybrid CNN Models}

The hybrid CNN–RNN architectures outperformed the purely recurrent models at both daily and weekly forecasting horizons. For the 24-hour horizon, the CNN–GRU model M5 with 256 hidden units achieved the best overall performance, with an RMSE of 0.132651, an $R^2$ of 0.668276, and a WQS of 0.827535. The CNN–LSTM model M6 with 64 hidden units followed closely, obtaining a WQS of 0.826709. The minimal difference of 0.000826 WQS units demonstrates that these two hybrid architectures delivered nearly equivalent next-day forecasting performance.

A similar pattern emerged for the 168-hour horizon (weekly forecasts). Model M5 with 1024 hidden units achieved a WQS of 0.782863, while M6 with 64 hidden units obtained 0.782017. Although M5 achieved the highest score at both horizons, the differences between the CNN–GRU and CNN–LSTM variants remained below 0.001 for WQS values. These findings indicate that the one-dimensional convolutional feature extractor contributed consistently to predictive performance, regardless of whether the recurrent component utilized a GRU or an LSTM. The overall results of the hybrid architectures are presented in Table~\ref{tab:hybrid_overall}.

\begin{table}[ht]
\centering
\caption{Test-set performance of the hybrid CNN-RNN  (GRU/LSTM) models for both forecasting horizons (daily/weekly)}
\label{tab:hybrid_overall}
\resizebox{1\linewidth}{!}{
\begin{tabular}{l r r r r r r r}
\hline
\textbf{Horizon} & \textbf{Model} & \textbf{Hidden} & \textbf{Layers} &
\textbf{RMSE}&
\textbf{R$^2$} &
\textbf{WQS} \\

\hline \hline
\multirow{2}{*}{\textbf{24 h}} 
& M5 &	256 &	1 &	\textbf{0.132651} & 	\textbf{0.668276} &	\textbf{0.827535} \\
& M6 &	64 &	1 &	0.133070 & 	0.665822 &	0.826709 \\

\hline
\multirow{2}{*}{\textbf{168 h}} 
& M5 &	1024 &	1 &	\textbf{0.156825} & 	\textbf{0.541613} &	\textbf{0.782863} \\
& M6 &	64 &	1 &	0.157282 & 	0.539212 &	0.782017 \\

\hline
\end{tabular}
}
\end{table}

Extending the forecasting horizon from 24 to 168 hours, corresponding to a shift from daily to weekly predictions, reduced the WQS of M5 and M6 by approximately 5.40\% for both the GRU and LSTM models. This decline aligns with the performance of purely recurrent models and reflects the increased uncertainty inherent in week-ahead forecasting. However, the hybrid architectures maintained relatively robust performance over the extended horizon, with both models achieving WQS values above 0.78. Variable-specific results indicate substantial differences in predictability among the meteorological quantities. For the 24-hour horizon, $ET_0$ and VPD were the most predictable variables, with maximum WQS values of 0.952746 and 0.951212, respectively. Wind speed exhibited a lower, yet still strong, maximum WQS of 0.810758. The sine and cosine components of wind direction were the most challenging targets, with maximum scores of 0.713504 and 0.711401, respectively.

For the 168-hour weekly forecast horizon, $ET_0$ remained the most predictable variable, with a maximum WQS of 0.936596, followed by VPD at 0.901950. Wind speed performance declined only moderately to approximately 0.782. In contrast, the transformed wind-direction components achieved WQS values between 0.636 and 0.658, indicating that directional variability was more challenging to forecast over a week. Variable-specific results for both horizons are presented in Table~\ref{tab:hybrid_variable}. The variable-wise scores of M5 and M6 were nearly identical, suggesting that most performance gains resulted from the convolutional feature extraction stage rather than the specific recurrent unit type. M5 performed marginally better for $ET_0$, VPD, and wind speed, while M6 achieved slightly higher scores for the transformed wind-direction components. However, these differences were insufficient to establish a clear overall advantage for either the GRU or LSTM recurrent backbone.

\begin{table}[ht]
\centering
\caption{Variable-specific WQS values of the hybrid CNN-RNN (GRU/LSTM) models for both forecasting horizons (daily/weekly).}
\label{tab:hybrid_variable}
\resizebox{1\linewidth}{!}{
\begin{tabular}{l r r r r r r}
\hline
\textbf{Horizon} & \textbf{Model} &
\textbf{WQS (ET0)} &
\textbf{WQS (VPD)} &
\textbf{WQS (Wind speed)} &
\textbf{WQS ($w_{sin}$) } &
\textbf{WQS ($w_{cos}$)} \\

\hline \hline
\multirow{2}{*}{\textbf{24 h}} 
& M5 & \textbf{0.952746}	& \textbf{0.951212} &	\textbf{0.810758} &	0.712908 &	0.710049 \\
& M6 & 0.952006 & 0.947614 &	0.809018 &	\textbf{0.713504} &	\textbf{0.711401} \\
\hline
\multirow{2}{*}{\textbf{168 h}}
& M5  & \textbf{0.936596} &	\textbf{0.901950} &	\textbf{0.781613} &	0.658067 &	0.636088 \\
& M6  &	0.935188 &	0.898974 &	0.781452 &	\textbf{0.658219} &	\textbf{0.636251} \\
\hline
\end{tabular}
}
\end{table}

Compared with the strongest purely recurrent configurations of GRU and LSTM models, the hybrid
models produced larger relative improvements for the 24-h horizon than for the 168-h horizon (weekly forecasts). At 24~h, M5 improved the WQS of the best GRU model of 256 units by 1.63\%, while M6 improved the best LSTM of 256 units result by 1.22\%. At 168~h, the corresponding gains decreased to 0.44\% and 0.45\%, respectively. This suggests that the one-dimensional convolutional layers were particularly effective at extracting local short-term patterns, whereas their relative benefit became smaller as the forecasting horizon increased. The direct comparison between the best recurrent and corresponding hybrid models is shown in Table~\ref{tab:hybrid_recurrent_comparison}.

\begin{table}[ht]
\centering
\caption{Comparison between the best purely recurrent models (GRU/LSTM) and their corresponding hybrid CNN-RNN variants (GRU/LSTM).}
\label{tab:hybrid_recurrent_comparison}
\resizebox{1\linewidth}{!}{
\begin{tabular}{l l r r r r r r}
\hline
\textbf{Horizon} & \textbf{Model Configuration} & \textbf{Recurrent's WQS} & \textbf{Hybrids's WQS} &
\textbf{Improvement}\\

\hline \hline
\multirow{2}{*}{\textbf{24 h}} 
& GRU (256 hidden units)  & 0.814284 & 0.827535 & +1.63\%\\
& LSTM (64 hidden units)  & 0.816755 & 0.826709 & +1.22\%\\

\hline
\multirow{2}{*}{\textbf{168 h}} 
& GRU (1024 hidden units) & 0.779465 & 0.782863 & +0.44\%\\
&  LSTM (64 hidden units) & 0.778542 & 0.782017 & +0.45\%\\

\hline
\end{tabular}
}
\end{table}

Overall, the results demonstrate that combining a one-dimensional convolutional feature extractor with recurrent sequence modelling provides a modest but consistent improvement over purely recurrent architectures. The benefit was most evident for the 24-h forecasting task and for variables exhibiting strong local and periodic temporal patterns, such as $ET_0$, VPD, and wind speed. Wind direction remained the most challenging target regardless of the recurrent backbone, indicating that improvements in its prediction may require additional enhancements in the hybrid modelling, or more spatiotemporal information used as input. The following subsection~\ref{sec:sum}, summarises this research findings.

\subsection{Results Summary}\label{sec:sum}

Across both forecasting horizons, daily and weekly, model depth had a stronger influence on performance than the choice between GRU and LSTM cells. The best recurrent results were consistently obtained by single-layer architectures, whereas
the four-layer models generally produced slightly lower scores and the ten-layer configurations deteriorated substantially. For the 24-h horizon, the best recurrent configurations were the single-layer GRU with 256 hidden units and the single-layer LSTM with 64 hidden units. Their WQS values differed by only 0.002471, with the LSTM achieving a relative advantage of approximately 0.30\%. For the 168-h horizon, the best configurations were the single-layer GRU with
1024 hidden units and the single-layer LSTM with 64 hidden units, whose WQS values differed by only 0.000923. These small differences indicate that neither recurrent cell type provided a consistent advantage across the two forecasting horizons. Increasing the hidden-state size beyond 64 units did not produce a significant improvement. Although the 256-unit GRU was the best recurrent GRU for the 24-h horizon and the 1024-unit GRU was the best for the 168-h horizon, the gains over the corresponding 64-unit configurations were small. For the LSTM models, 64 hidden units produced the best result
at both forecasting horizons, while the 256- and 1024-unit variants provided similar or slightly lower performance. 

When GRU and LSTM models with the same hidden-state size were compared, the LSTM achieved slightly higher WQS values for all tested configurations at the 24-h horizon, with relative improvements ranging from approximately
0.19\% to 0.73\%. At the 168-h horizon, the LSTM also outperformed the GRU for hidden sizes of 64, and 256 units, with gains ranging from 0.54\% to 1.50\%. The only exception was the 1024-unit configuration, for which
the GRU achieved a 0.75\% higher WQS. Overall, the differences were modest, indicating that neither recurrent cell type provided a decisive advantage, with more units needed for a GRU model to reach the LSTM performance. The best-performing GRU configurations required substantially larger hidden-state dimensions than the best LSTM configurations. Specifically,
the GRU used 300\% more hidden units for the 24-h horizon (256 versus 64 units) and 1500\% more for the 168-h horizon
(1024 versus 64 units), corresponding to four and sixteen times as many hidden units, respectively. By contrast, the ten-layer models reduced the best recurrent WQS by approximately 23.7\% for the 24-h horizon and 10.3\% for the 168-h horizon, while also requiring considerably more trainable parameters and longer training times.

The results also revealed a clear dependence on both the predicted meteorological variable and the recurrent-model configuration. For the 24-h horizon, reference evapotranspiration and vapour pressure deficit were the most predictable variables. The 64-unit LSTM achieved the highest $ET_0$ WQS of 0.948005, while the 256-unit LSTM produced the best VPD score of 0.940687. The 256-unit LSTM also achieved the highest wind-speed WQS of 0.799244, although its advantage over the 64-unit LSTM was negligible. For the wind-direction components, the 64-unit LSTM performed best, with WQS values of 0.699428 for $w_{\sin}$ and 0.696642 for $w_{\cos}$. In contrast, the 1024-unit GRU did not provide an advantage at the 24-h horizon and produced lower scores than the 64- and 256-unit LSTM configurations for most variables.

For the 168-h horizon, the 1024-unit GRU achieved the best performance for most target variables, obtaining WQS values of 0.932572 for $ET_0$, 0.778015 for wind speed, 0.655488 for $w_{\sin}$, and 0.632864 for $w_{\cos}$. The only exception was VPD, for which the 256-unit LSTM achieved the highest WQS of 0.899691. Nevertheless, $ET_0$ and VPD remained the most predictable variables, whereas the sine and cosine components of wind direction exhibited the lowest performance. The 1024-unit LSTM generally performed worse than the 64- and 256-unit LSTM configurations at the weekly horizon, indicating that increasing the hidden-state size did not consistently improve LSTM performance. Overall, the results suggest that moderate LSTM configurations were more effective for the 24-h forecasts, while the larger 1024-unit GRU was more competitive for the 168-h forecasting task.

The hybrid CNN-RNN architectures produced the strongest overall results. For the 24-h horizon, the CNN-GRU model M5 achieved a WQS of 0.827535, while the CNN-LSTM model M6 obtained 0.826709. These values represent relative improvements of 1.63\% and 1.22\% over the best purely recurrent GRU and LSTM configurations, respectively. For the 168-h horizon, the
corresponding improvements decreased to 0.44\% for M5 and 0.45\% for M6. This pattern indicates that one-dimensional convolutional feature extraction was particularly useful for identifying local short-term dependencies, while its relative contribution became smaller as the forecasting horizon increased. The nearly identical results of M5 and M6 also suggest that the main benefit originated from the convolutional front end rather than from the specific recurrent backbone.

Overall, the experiments demonstrate that shallow, single-layer recurrent architectures provide a better balance between predictive accuracy and model complexity than very deep recurrent networks. Among the purely recurrent models, the best 24-h results were obtained by the 64-unit LSTM and the 256-unit GRU, with WQS values of 0.816755 and 0.814284, respectively. For the 168-h horizon, the strongest recurrent configurations were the 1024-unit GRU and the 64-unit LSTM, which achieved WQS values of 0.779465 and 0.778542. Adding a one-dimensional convolutional feature extractor produced a small but consistent improvement over these best recurrent baselines. Specifically, the CNN-GRU model increased the WQS to 0.827535 at 24~h and 0.782863 at 168~h, corresponding to relative improvements of 1.63\% and 0.44\% over the best GRU configurations. Similarly, the CNN-LSTM model achieved WQS values of 0.826709 and 0.782017, improving upon the best LSTM models by 1.22\% and 0.45\%, respectively.

When the hybrid CNN-GRU and CNN-LSTM architectures were compared, the CNN-GRU model M5 achieved marginally higher overall WQS values at both forecasting horizons. At the 24-h horizon, M5 with 256 hidden units achieved a WQS of 0.827535, compared with 0.826709 for M6 with 64 hidden units, corresponding to a relative difference of only 0.10\%. At the 168-h horizon, M5 with 1024 hidden units achieved a WQS of 0.782863, whereas M6 retained 64 hidden units and obtained 0.782017, a difference of approximately 0.11\%. These comparisons were not performed at equal hidden-state sizes: the CNN-GRU used 300\% more hidden units than the CNN-LSTM at the 24-h horizon and 1500\% more at the 168-h horizon, corresponding to four and sixteen times as many units, respectively. Therefore, although the CNN-GRU produced the highest absolute scores, the CNN-LSTM achieved practically equivalent performance with substantially lower recurrent capacity. Relative to their strongest purely recurrent counterparts, the CNN-GRU improved the best GRU WQS by 1.63\% at 24~h and 0.44\% at 168~h, while the CNN-LSTM improved the best LSTM results by 1.22\% and 0.45\%, respectively. These findings indicate that the main performance gain originated from the convolutional feature extractor, whereas the choice of the recurrent backbone had only a limited effect.

The variable-specific results also depended on the hybrid recurrent backbone. For the 24-h horizon, the CNN-GRU model M5 achieved the highest WQS values for $ET_0$, VPD, and wind speed, with scores of 0.952746, 0.951212, and 0.810758, respectively. Compared with the best purely recurrent configurations, these values corresponded to relative improvements of approximately 0.50\% for $ET_0$, 1.12\% for VPD, and 1.44\% for wind speed. In contrast, the CNN-LSTM model M6 achieved the best results for the wind-direction components, with WQS values of 0.713504 for $w_{\sin}$ and 0.711401 for $w_{\cos}$, improving upon the best recurrent scores by approximately 2.01\% and 2.12\%, respectively. At the 168-h horizon, M5 remained the strongest model for $ET_0$, VPD, and wind speed, achieving WQS values of 0.936596, 0.901950, and 0.781613, whereas M6 again produced the highest scores for $w_{\sin}$ and $w_{\cos}$, with values of 0.658219 and 0.636251. The corresponding gains over the best purely recurrent models were smaller, ranging from approximately 0.25\% to 0.54\%. Overall, $ET_0$ and VPD remained the most predictable variables, while the transformed wind-direction components exhibited the lowest performance.

\begin{figure}[ht]
	\centering
	\includegraphics[width=\linewidth]{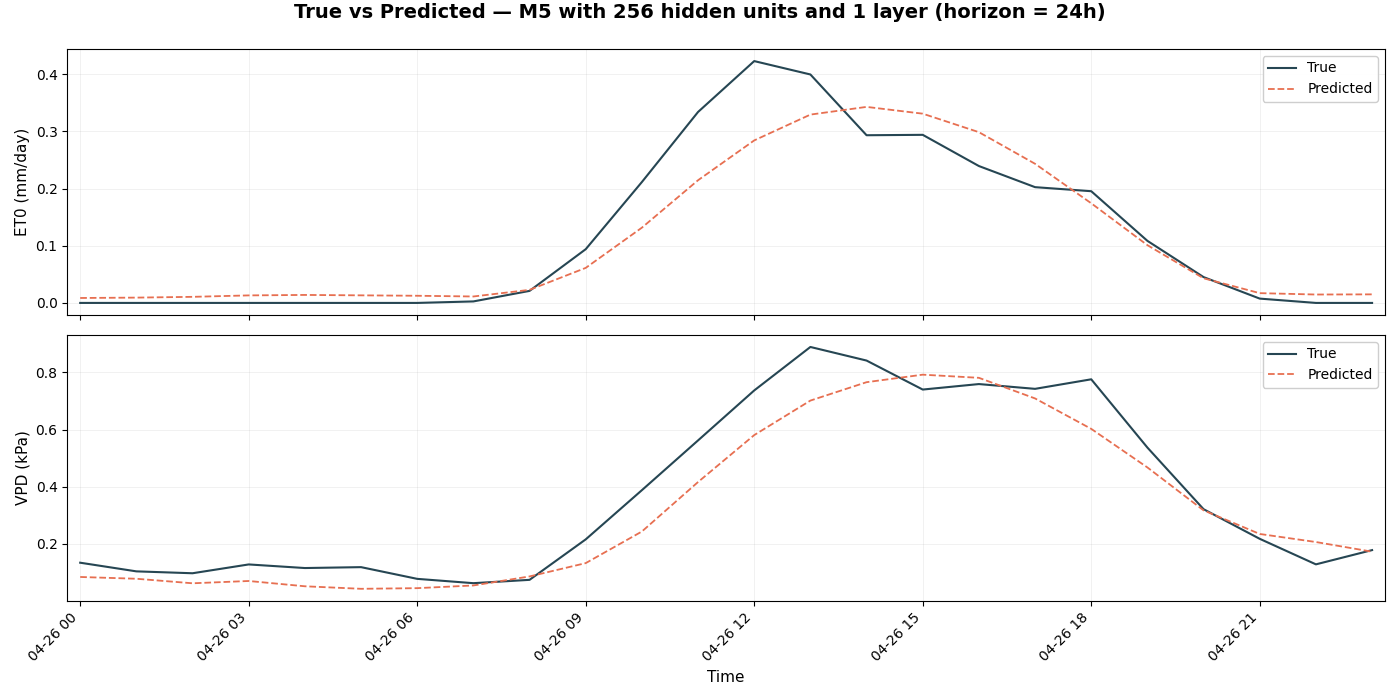}
	\caption{Observed and predicted reference evapotranspiration ($ET_0$)
		and vapour pressure deficit (VPD) values produced by the
		best-performing model for the 24-h forecasting horizon.}
	\label{fig:day_predictions}
\end{figure}

\begin{figure}[ht]
	\centering
	\includegraphics[width=\linewidth]{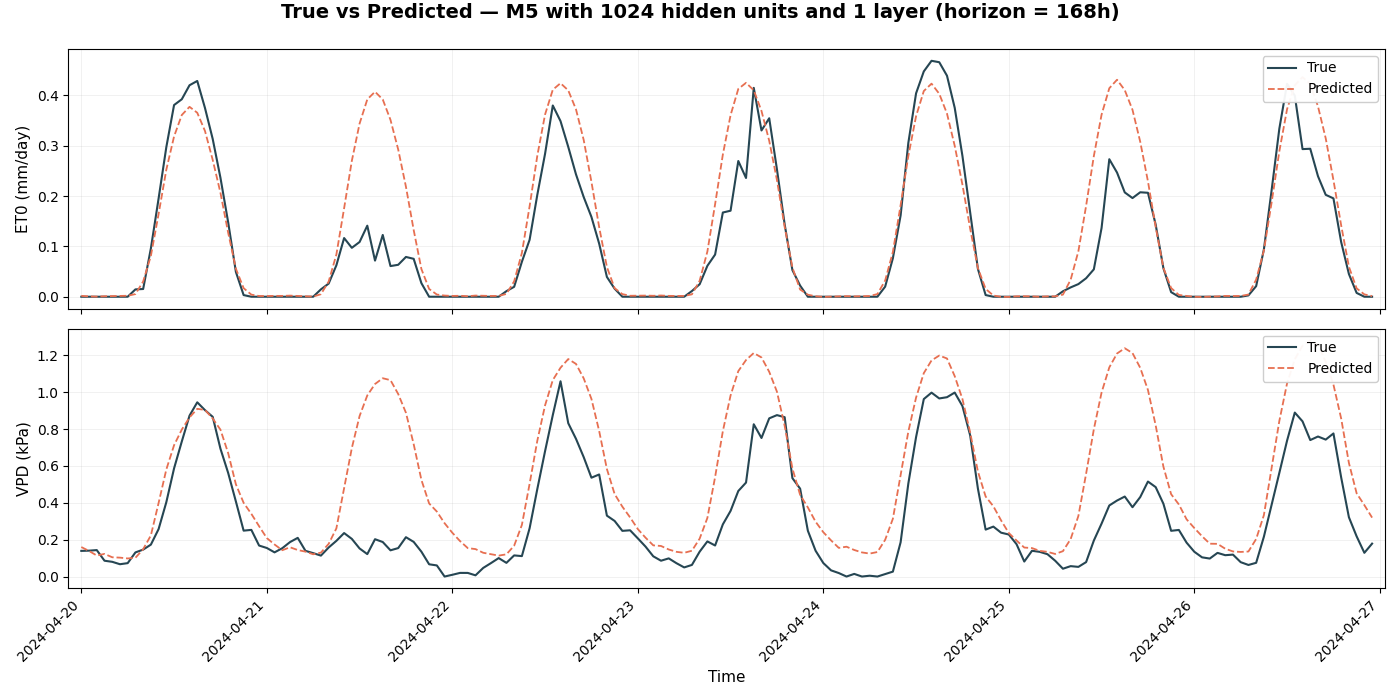}
	\caption{Observed and predicted reference evapotranspiration ($ET_0$)
		and vapour pressure deficit (VPD) values produced by the
		best-performing model for the 168-h forecasting horizon.}
	\label{fig:week_predictions}
\end{figure}

Figure~\ref{fig:day_predictions} presents the observed and predicted $ET_0$ and VPD values generated by the best-performing model for the 24-h forecasting horizon, while Figure~\ref{fig:week_predictions} provides the corresponding comparison for the 168-h forecasting horizon, illustrating the more demanding week-ahead prediction task.

\section{Conclusions}
This study provides a comparative evaluation of GRU, LSTM, CNN-GRU, and CNN-LSTM architectures for multivariate meteorological forecasting using hourly ERA5 data for Ioannina, Greece. The models jointly predict reference evapotranspiration ($ET_0$), vapour pressure deficit (VPD), wind speed, and the sine and cosine components of wind direction over two forecasting horizons: 24 hours and 168 hours. The experiments assess the effects of recurrent cell type, hidden-state size, network depth, convolutional feature extraction, forecasting horizon, and target-variable characteristics. The main conclusions are as follows:
\begin{itemize}
	\item \textbf{Single-layer recurrent models provided the strongest purely recurrent performance.}
	At the 24-hour horizon, the optimal recurrent configurations were the 64-unit LSTM and the 256-unit GRU, achieving WQS values of 0.816755 and 0.814284, respectively. For the 168-hour horizon, the 1024-unit GRU and the 64-unit LSTM yielded the highest WQS values of 0.779465 and 0.778542. These findings indicate that shallow recurrent architectures outperformed deeper stacked alternatives across both forecasting tasks.
	
	\item \textbf{LSTM and GRU achieved similar predictive accuracy, but the LSTM generally required fewer hidden units.}
	Comparisons of models with identical hidden-state sizes revealed that the LSTM consistently achieved marginally higher WQS values at the 24-hour horizon. At the 168-hour horizon, the LSTM outperformed the GRU for 64 and 256 hidden units, while only the 1024-unit GRU slightly exceeded the corresponding LSTM. Notably, the best-performing GRU required 256 hidden units at 24 hours and 1024 units at 168 hours, whereas the optimal LSTM utilized only 64 units at both horizons. Consequently, the best GRU models required four times as many hidden units at 24 hours and sixteen times as many at 168 hours.
	
	\item \textbf{Increasing hidden-state size  did not provide a consistent benefit.}
	The LSTM achieved optimal performance with 64 hidden units at both horizons, whereas its 256- and 1024-unit variants yielded similar or slightly reduced scores. Larger GRU models sometimes improved performance, especially at the 168-hour horizon, but the improvements over the 64-unit GRU were minimal relative to the increase in model capacity. Therefore, hidden-state size should be selected based on the recurrent architecture and forecasting horizon, rather than increased indiscriminately.
	
	\item \textbf{Excessive recurrent depth by stacking layers reduced both accuracy and efficiency.}
	The ten-layer GRU and LSTM configurations reduced the best recurrent WQS by approximately 23.7\% at the 24-hour horizon and 10.3\% at the 168-hour horizon. These models also contained tens of millions of parameters and required significantly longer training times. Under the evaluated conditions, increased depth led to optimization and generalization challenges without enhancing forecast accuracy.
	
	\item \textbf{The hybrid CNN-RNN architectures achieved the highest overall accuracy.}
	The CNN-GRU model achieved the highest absolute WQS values, reaching 0.827535 at 24 hours and 0.782863 at 168 hours. These results represent relative improvements of 1.63\% and 0.44\% over the strongest purely recurrent GRU configurations, respectively. The CNN-LSTM model produced comparable WQS values of 0.826709 and 0.782017, surpassing the best purely recurrent LSTM models by 1.22\% and 0.45\%, respectively. The greater improvement at the 24-hour horizon suggests that convolutional feature extraction is particularly effective for capturing local short-term temporal patterns. Among the hybrid configurations tested, the CNN-GRU model outperformed the CNN-LSTM model by approximately 0.10\% at 24 hours and 0.11\% at 168 hours. However, this difference cannot be attributed solely to the recurrent cell type, as the CNN-GRU and CNN-LSTM models used different hidden-state sizes while maintaining the same single-layer recurrent depth. Specifically, at the 24-hour horizon, the CNN-GRU model used 256 hidden units, whereas the CNN-LSTM model used 64. At the 168-hour horizon, the CNN-GRU model used 1024 hidden units, while the CNN-LSTM again used 64. Therefore, the minor performance difference is not solely due to the choice between GRU and LSTM cells.
	
	\item \textbf{The effectiveness of the models depended strongly on the predicted meteorological variable.}
	Reference evapotranspiration and vapor pressure deficit (VPD) were the most predictable variables. The highest hybrid model scores reached 0.952746 for $ET_0$ and 0.951212 for VPD at the 24-hour horizon, with corresponding 168-hour scores remaining high at 0.936596 and 0.901950. Wind speed proved more challenging to forecast, as its best WQS decreased from 0.810758 at
	24~h to 0.781613 at 168~h.
	
	\item \textbf{Wind direction remained the most difficult forecasting target.}
	The highest 24-hour WQS values were 0.713504 for $w_{\sin}$ and 0.711401 for $w_{\cos}$. At the 168-hour horizon, these values declined to 0.658219 and 0.636251, respectively. The cosine component was approximately 0.3\% lower than the sine component at 24 hours, with the difference increasing to about 3.3\% at 168 hours. These findings suggest that the irregular and circular characteristics of wind direction were not fully captured by the evaluated architectures.
	
	\item \textbf{The reduction in accuracy with increasing forecast horizon was variable-dependent.}
	Comparison of the best hybrid model results showed that WQS decreased by approximately 1.7\% for $ET_0$, 5.2\% for VPD, 3.6\% for wind speed, 7.8\% for $w_{\sin}$, and 10.6\% for $w_{\cos}$ when extending the forecast horizon from one day to one week. This indicates that reference evapotranspiration was least affected by the longer horizon, while the cosine representation of wind direction was most affected.
\end{itemize}

Several limitations are present in this study. The analysis utilized meteorological data from a single ERA5 grid point near the University of Ioannina (city of Ioannina), which precluded examination of spatial dependencies among multiple stations. The multilayer models were assessed using a representative subset of possible depth-width combinations, rather than an exhaustive parameter grid. Additionally, the learning rate, dropout probability, batch size, optimizer, and stopping criteria were held constant across all architectures, and systematic hyperparameter optimization was not conducted. Consequently, the conclusions are specific to the dataset, forecasting horizons, variables, and model configurations examined.

Future research should evaluate the models using observations from multiple meteorological stations and additional geographical regions. Attention-based architectures, such as transformers and temporal fusion transformers, should be systematically compared with recurrent and hybrid models. Incorporating additional temporal features, including hour of day, day of year, season, and cyclic calendar encodings, may enhance the representation of periodic behavior. Further studies should also implement systematic hyperparameter optimization and develop probabilistic forecasts with uncertainty intervals specifically tailored for wind-direction prediction.

\subsection*{Models and code availability}
The code is publicly available at: \url{https://sensors.math.uoi.gr:3002/MCSL_Team/AgriPlan}.

{
    \small
    \bibliographystyle{IEEEtranS}
    \bibliography{main}

@misc{lam2023graphcast,
	title={GraphCast: Learning skillful medium-range global weather forecasting}, 
	author={Remi Lam and Alvaro Sanchez-Gonzalez and Matthew Willson and Peter Wirnsberger and Meire Fortunato and Ferran Alet and Suman Ravuri and Timo Ewalds and Zach Eaton-Rosen and Weihua Hu and Alexander Merose and Stephan Hoyer and George Holland and Oriol Vinyals and Jacklynn Stott and Alexander Pritzel and Shakir Mohamed and Peter Battaglia},
	year={2023},
	eprint={2212.12794},
	archivePrefix={arXiv},
	primaryClass={cs.LG},
	url={https://arxiv.org/abs/2212.12794}
}

@misc{zippenfenig2023openmeteo,
	author       = {Zippenfenig, Patrick},
	title        = {{Open-Meteo.com Weather API}},
	year         = {2023},
	howpublished = {Zenodo},
	doi          = {10.5281/zenodo.7970649},
	url          = {https://doi.org/10.5281/zenodo.7970649},
	note         = {Available at: \url{https://open-meteo.com/}
	(accessed 11 Apr 2024).}
}

@misc{hersbach2023era5,
	author       = {Hersbach, H. and Bell, B. and Berrisford, P. and
	Biavati, G. and Horányi, A. and Muñoz Sabater, J. and
	Nicolas, J. and Peubey, C. and Radu, R. and Rozum, I. and
	Schepers, D. and Simmons, A. and Soci, C. and Dee, D. and
	Thépaut, J.-N.},
	title        = {{ERA5} hourly data on single levels from 1940 to present},
	year         = {2023},
	howpublished = {ECMWF Copernicus Climate Data Store},
	publisher    = {ECMWF},
	doi          = {10.24381/cds.adbb2d47},
	url          = {https://doi.org/10.24381/cds.adbb2d47},
	note         = {Available at:
	\url{https://cds.climate.copernicus.eu/doi/10.24381/cds.adbb2d47}
	(accessed 08 Apr 2024)}
}

@article{hochreiter1997lstm,
	author = {Hochreiter, Sepp and Schmidhuber, Jürgen},
	year = {1997},
	month = {11},
	pages = {1735-1780},
	title = {Long Short-Term Memory},
	volume = {9},
	journal = {Neural Computation},
	doi = {10.1162/neco.1997.9.8.1735}
}

@article{gers2000lstm,
	author = {Gers, Felix and Schmidhuber, Jürgen and Cummins, Fred},
	year = {2000},
	month = {10},
	pages = {2451-2471},
	title = {Learning to Forget: Continual Prediction with LSTM},
	volume = {12},
	journal = {Neural Computation},
	doi = {10.1162/089976600300015015}
}

@misc{chung2014gru,
	author        = {Chung, Junyoung and Gulcehre, Caglar and
	Cho, KyungHyun and Bengio, Yoshua},
	title         = {Empirical Evaluation of Gated Recurrent Neural
	Networks on Sequence Modeling},
	year          = {2014},
	howpublished  = {arXiv preprint},
	eprint        = {1412.3555},
	archiveprefix = {arXiv},
	primaryclass  = {cs.NE},
	doi           = {10.48550/arXiv.1412.3555},
	url           = {https://doi.org/10.48550/arXiv.1412.3555},
	note          = {Available at:
	\url{https://arxiv.org/abs/1412.3555}
	(accessed 25 Mar 2021)}
}

@article{ige2024cnn1d,
	author={Ige, Ayokunle Olalekan and Sibiya, Malusi},
	journal={IEEE Access}, 
	title={State-of-the-Art in 1D Convolutional Neural Networks: A Survey}, 
	year={2024},
	volume={12},
	pages={144397-144419},
	doi={10.1109/ACCESS.2024.3433513}
}

@mastersthesis{fischer2025comparison,
	author = {Fischer, Carmen},
	title = {Comparison of Machine Learning Algorithms},
	school = {FH Campus Wien},
	year = {2025},
	type = {Bachelor's thesis},
	url = {https://pub.hcw.ac.at/obvfcwhs/content/titleinfo/12219798}
}

@article{domingos2025exploring,
	title = {Exploring Machine Learning, Deep Learning, and Explainable AI Methods for Seasonal Precipitation Prediction in South America},
	author = {Matheus Corr{\^e}a Domingos and Valdivino Alexandre de Santiago J{\'u}nior and Juliana Aparecida Anochi and Elcio Hideiti Shiguemori and Lu{\'i}sa Mirelle Costa dos Santos and H{\'e}rcules Carlos dos Santos Pereira and Andr{\'e} Estevam Costa Oliveira},
	year = {2025},
	eprint = {2512.13910},
	archivePrefix = {arXiv},
	primaryClass = {cs.LG},
	url = {https://arxiv.org/abs/2512.13910},
	journal = {arXiv preprint},
	note = {Submitted for publication}
}

@misc{paszke2019pytorch,
	author        = {Paszke, Adam and Gross, Sam and Massa, Francisco and
	Lerer, Adam and Bradbury, James and Chanan, Gregory and
	Killeen, Trevor and Lin, Zeming and Gimelshein, Natalia and
	Antiga, Luca and Desmaison, Alban and K{\"o}pf, Andreas and
	Yang, Edward and DeVito, Zach and Raison, Martin and
	Tejani, Alykhan and Chilamkurthy, Sasank and
	Steiner, Benoit and Fang, Lu and Bai, Junjie and
	Chintala, Soumith},
	title         = {{PyTorch}: An Imperative Style, High-Performance
	Deep Learning Library},
	year          = {2019},
	howpublished  = {arXiv preprint},
	eprint        = {1912.01703},
	archiveprefix = {arXiv},
	primaryclass  = {cs.LG},
	doi           = {10.48550/arXiv.1912.01703},
	url           = {https://doi.org/10.48550/arXiv.1912.01703},
	note          = {Available at:
	\url{https://arxiv.org/abs/1912.01703}
	(accessed 15 Apr 2020)}
}

@misc{kingma2017adam,
	author        = {Kingma, Diederik P. and Ba, Jimmy},
	title         = {{Adam}: A Method for Stochastic Optimization},
	year          = {2017},
	howpublished  = {arXiv preprint},
	eprint        = {1412.6980},
	archiveprefix = {arXiv},
	primaryclass  = {cs.LG},
	doi           = {10.48550/arXiv.1412.6980},
	url           = {https://doi.org/10.48550/arXiv.1412.6980},
	note          = {Available at:
	\url{https://arxiv.org/abs/1412.6980}
	(accessed 15 Apr 2020)}
}

@article{hersbach2020era5,
	author = {Hersbach, Hans and Bell, Bill and Berrisford, Paul and Hirahara, Shoji and Horányi, András and Muñoz-Sabater, Joaquín and Nicolas, Julien and Peubey, Carole and Radu, Raluca and Schepers, Dinand and Simmons, Adrian and Soci, Cornel and Abdalla, Saleh and Abellan, Xavier and Balsamo, Gianpaolo and Bechtold, Peter and Biavati, Gionata and Bidlot, Jean and Bonavita, Massimo and De Chiara, Giovanna and Dahlgren, Per and Dee, Dick and Diamantakis, Michail and Dragani, Rossana and Flemming, Johannes and Forbes, Richard and Fuentes, Manuel and Geer, Alan and Haimberger, Leo and Healy, Sean and Hogan, Robin J. and Hólm, Elías and Janisková, Marta and Keeley, Sarah and Laloyaux, Patrick and Lopez, Philippe and Lupu, Cristina and Radnoti, Gabor and de Rosnay, Patricia and Rozum, Iryna and Vamborg, Freja and Villaume, Sebastien and Thépaut, Jean-Noël},
	title = {The ERA5 global reanalysis},
	journal = {Quarterly Journal of the Royal Meteorological Society},
	volume = {146},
	number = {730},
	pages = {1999-2049},
	doi = {10.1002/qj.3803},
	url = {https://rmets.onlinelibrary.wiley.com/doi/abs/10.1002/qj.3803},
	year = {2020}
}

@article{scikit-learn,
	title={Scikit-learn: Machine Learning in {P}ython},
	author={Pedregosa, F. and Varoquaux, G. and Gramfort, A. and Michel, V. and Thirion, B. and Grisel, O. and Blondel, M. and Prettenhofer, P. and Weiss, R. and Dubourg, V. and Vanderplas, J. and Passos, A. and Cournapeau, D. and Brucher, M. and Perrot, M. and Duchesnay, E.},
	journal={Journal of Machine Learning Research},
	volume={12},
	pages={2825--2830},
	year={2011}
}

@book{allen1998crop,
	author    = {Allen, Richard G. and Pereira, Luis S. and Raes, Dirk
	and Smith, Martin},
	title     = {Crop Evapotranspiration: Guidelines for Computing
	Crop Water Requirements},
	series    = {FAO Irrigation and Drainage Paper},
	number    = {56},
	publisher = {Food and Agriculture Organization of the United Nations},
	address   = {Rome, Italy},
	year      = {1998},
	isbn      = {978-92-5-104219-9}
}

@article{tsolaki26,
	author = {Tsolaki, Christina and Kokkonis, George and Valsamidis, Stavros and Kontogiannis, Sotirios},
	title = {Water Quality Identification: Integrating IoT Sensors and Deep Learning for Near-Real-Time Water Quality Assessment},
	journal = {Applied Sciences},
	volume = {16},
	year = {2026},
	number = {10},
	article-number= {4868},
	url = {https://www.mdpi.com/2076-3417/16/10/4868},
	issn = {2076-3417},
	doi = {10.3390/app16104868}
}

@article{srivastava2014dropout,
	author  = {Srivastava, Nitish and Hinton, Geoffrey and
	Krizhevsky, Alex and Sutskever, Ilya and
	Salakhutdinov, Ruslan},
	title   = {Dropout: A Simple Way to Prevent Neural Networks
	from Overfitting},
	journal = {Journal of Machine Learning Research},
	year    = {2014},
	volume  = {15},
	number  = {56},
	pages   = {1929--1958},
	url     = {https://jmlr.org/papers/v15/srivastava14a.html}
}

@inproceedings{ioffe2015batchnorm,
	author    = {Ioffe, Sergey and Szegedy, Christian},
	title     = {Batch Normalization: Accelerating Deep Network Training
	by Reducing Internal Covariate Shift},
	booktitle = {Proceedings of the 32nd International Conference
	on Machine Learning},
	series    = {Proceedings of Machine Learning Research},
	volume    = {37},
	pages     = {448--456},
	year      = {2015},
	publisher = {PMLR},
	url       = {https://proceedings.mlr.press/v37/ioffe15.html}
}

@phdthesis{kontogiannis2026thingsai,
	author      = {Kontogiannis, Sotirios},
	title       = {Distributed Systems and Algorithms for Measurement
	Collection, Decision Making, and Visualization of
	Georeferenced Information with Applications in
	Viticulture},
	school      = {Aristotle University of Thessaloniki},
	year        = {2026},
	type        = {Doctoral dissertation},
	doi         = {10.26262/heal.auth.ir.368402},
	url         = {https://doi.org/10.26262/heal.auth.ir.368402},
	note        = {Available at:
	\url{https://ikee.lib.auth.gr/record/368402/?ln=el}
	(accessed 11 Mar 2026)}
}

@article{roy2022et0,
	author  = {Roy, Dilip Kumar and Sarkar, Tapash Kumar and
	Kamar, Sheikh Shamshul Alam and Goswami, Torsha and
	Muktadir, Md Abdul and Al-Ghobari, Hussein M. and
	Alataway, Abed and Dewidar, Ahmed Z. and
	El-Shafei, Ahmed A. and Mattar, Mohamed A.},
	title   = {Daily Prediction and Multi-Step Forward Forecasting of
	Reference Evapotranspiration Using {LSTM} and
	{Bi-LSTM} Models},
	journal = {Agronomy},
	year    = {2022},
	volume  = {12},
	number  = {3},
	pages   = {594},
	doi     = {10.3390/agronomy12030594},
	url     = {https://doi.org/10.3390/agronomy12030594}
}

@article{sarkar2025et0,
	author  = {Sarkar, Suman Saurabh and Bedi, Jatin and Jain, Sushma},
	title   = {A Deep Learning Based Framework for Enhanced Reference
	Evapotranspiration Estimation: Evaluating Accuracy and
	Forecasting Strategies},
	journal = {Scientific Reports},
	year    = {2025},
	volume  = {15},
	pages   = {15136},
	doi     = {10.1038/s41598-025-99713-2},
	url     = {https://doi.org/10.1038/s41598-025-99713-2}
}

@article{elbeltagi2023vpd,
	author  = {Elbeltagi, Ahmed and Srivastava, Aman and Deng, Jinsong and
	Li, Zhibin and Raza, Ali and Khadke, Leena and
	Yu, Zhoulu and El-Rawy, Mustafa},
	title   = {Forecasting Vapor Pressure Deficit for Agricultural Water
	Management Using Machine Learning in Semi-Arid
	Environments},
	journal = {Agricultural Water Management},
	year    = {2023},
	volume  = {283},
	pages   = {108302},
	doi     = {10.1016/j.agwat.2023.108302},
	url     = {https://doi.org/10.1016/j.agwat.2023.108302}
}

@article{vurro2019vpd,
	author  = {Vurro, Filippo and Janni, Michela and Copped{\`e}, Nicola and
	Gentile, Francesco and Manfredi, Riccardo and
	Bettelli, Manuele and Zappettini, Andrea},
	title   = {Development of an In Vivo Sensor to Monitor the Effects of
	Vapour Pressure Deficit ({VPD}) Changes to Improve Water
	Productivity in Agriculture},
	journal = {Sensors},
	year    = {2019},
	volume  = {19},
	number  = {21},
	pages   = {4667},
	doi     = {10.3390/s19214667},
	url     = {https://doi.org/10.3390/s19214667}
}
}


\end{document}